\documentclass[conference]{IEEEtran}
\IEEEoverridecommandlockouts
\usepackage{cite}
\usepackage{amsmath,amssymb,amsfonts}
\usepackage{algorithmic}
\usepackage{graphicx}
\usepackage{textcomp}
\usepackage{xcolor}
\usepackage{epsfig}
\usepackage{multirow}

\def\BibTeX{{\rm B\kern-.05em{\sc i\kern-.025em b}\kern-.08em
    T\kern-.1667em\lower.7ex\hbox{E}\kern-.125emX}}

\begin{document}
\IEEEoverridecommandlockouts
\IEEEpubid{\makebox[\columnwidth]{\textbf{\emph{ACCEPTED BY IEEE MMSP 2020}} \hfill} \hspace{\columnsep}\makebox[\columnwidth]{ }}
\title{Detection of Gait Abnormalities caused by Neurological Disorders}
\author{\IEEEauthorblockN{Daksh Goyal}
\IEEEauthorblockA{\textit{Department of Civil Engineering} \\
\textit{NIT Karnataka}\\
Surathkal, India \\
dakshgoyal.171cv111@nitk.edu.in}
\and
\IEEEauthorblockN{Koteswar Rao Jerripothula}
\IEEEauthorblockA{\textit{Department of Computer Science and Engineering} \\
\textit{Indraprastha Institute of Information Technology Delhi}\\
New Delhi, India \\
koteswar@iiitd.ac.in}
\and
\IEEEauthorblockN{Ankush Mittal}
\IEEEauthorblockA{
\textit{Raman Classes}\\
Roorkee, India \\
dr.ankush.mittal@gmail.com}
}

\maketitle

\begin{abstract}
In this paper, we leverage gait to potentially detect some of the important neurological disorders, namely Parkinson's disease, Diplegia, Hemiplegia, and Huntington's Chorea. Persons with these neurological disorders often have a very abnormal gait, which motivates us to target gait for their potential detection. Some of the abnormalities involve the circumduction of legs, forward-bending, involuntary movements, etc. To detect such abnormalities in gait, we develop gait features from the key-points of the human pose, namely shoulders, elbows, hips, knees, ankles, etc. To evaluate the effectiveness of our gait features in detecting the abnormalities related to these diseases, we build a synthetic video dataset of persons mimicking the gait of persons with such disorders, considering the difficulty in finding a sufficient number of people with these disorders. We name it \textit{NeuroSynGait} video dataset. Experiments demonstrated that our gait features were indeed successful in detecting these abnormalities. 
\end{abstract}


\section{Introduction}

A neurological disorder\cite{Yolcu2019} is a term given to certain abnormalities generated due to the malfunctioning of the human body's nervous system. The human body's nervous system essentially consists of the brain, the nerves, and the spinal cords. The nerves are responsible for connecting the brain and the spinal cord and mutually connecting these to the various parts of the body. It controls and coordinates even different bodily voluntary and involuntary actions. The nervous system is quite sophisticated and has an excellent system of information transfer. Therefore, any fault or disorder in this system can cause trouble in performing even simple movements such as writing, walking, and eating correctly. These troubles lead to various visual abnormalities, which can be recorded in videos \cite{10.1007/978-3-319-46478-7_12} and analyzed. We choose to analyze gait (a person's manner of walking) to detect these abnormalities and, thus, potentially even the neurological disorder. Considering this as a video analytics problem, we can employ different computer vision and machine learning algorithms to build a computer-assisted diagnosis system. Given a video of the subject's gait \cite{zhang2013score,gianaria2013gait,dao2015kinect,khan2019generic}, our primary goal is to be able to detect different gait abnormalities related to neurological disorders. 

\begin{figure}
        \begin{center}
            \includegraphics[width=1\linewidth]{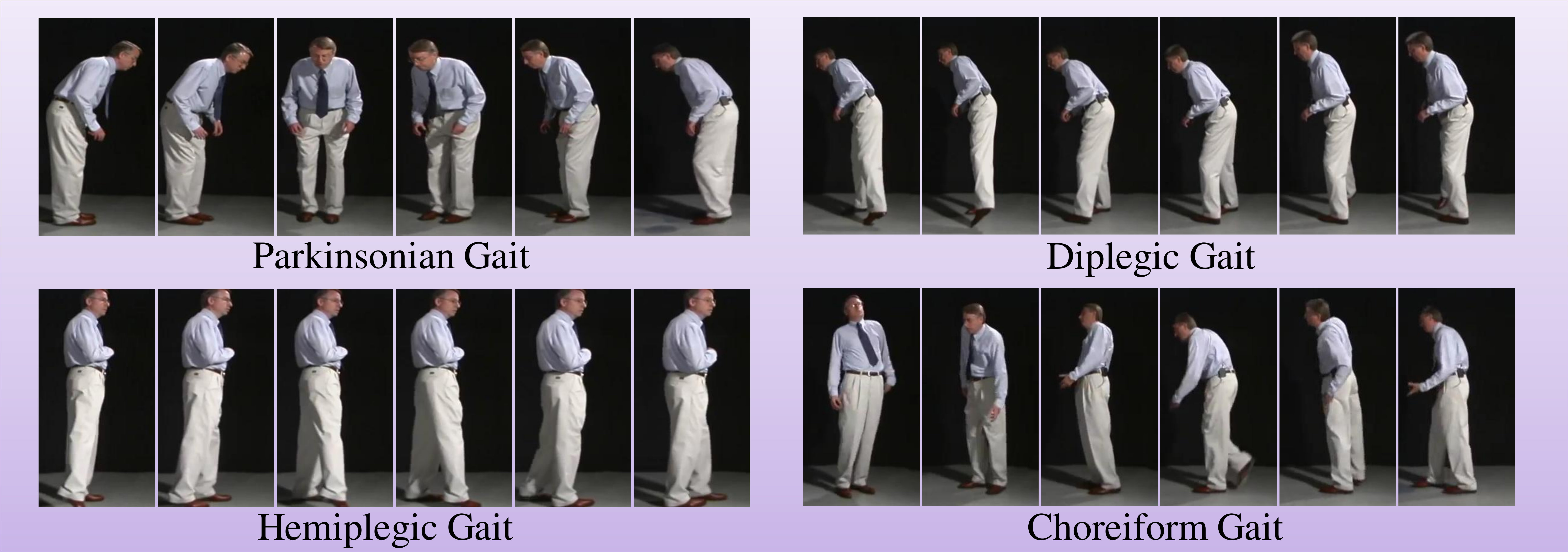}
        \end{center}
        \vspace{-4mm}
        \caption{Gait Abnormalities }\label{fig:ga}
\end{figure}

In some of the neurological disorders, gait-related abnormalities are excellent visible symptoms to detect those disorders. Even doctors consider such visible symptoms as a basis for their diagnosis to identify the root cause of the problem. This research attempts to expedite the diagnosis process by automating this step with computer vision and machine learning algorithms. It is not only for saving time (or faster diagnosis) but also because it is much cheaper than the existing electronic instruments based diagnosis. Also, special assistance could be given to such people at public places and events by automatically identifying them. Parkinson's disease can be identified by what is known as parkinsonian gait \cite{nandy2019statistical}, where a person bends forward and walks very slowly. Similarly, Hemiplegia can be identified by hemiplegic gait, characterized by leg's circumduction (conical movement). In Diplegia, the gait is known as diplegic gait, which involves circumduction (conical movement) of both the legs. Lastly, Huntington's Chorea is characterized by its choreaform gait that involves completely involuntary and unbalanced walking. Sample cropped frames of a demo of all these abnormal gaits are given in Fig.~\ref{fig:ga}. These observations could be modeled into video-based predictors (features) using state-of-the-art pose-estimation algorithms (of computer vision). These features can subsequently be employed for detecting the gait abnormalities through machine learning algorithms.

That said, there are quite a few challenges involved. First, there are no publicly available gait video datasets for neurological disorders. Second, there are no readily available video features to quantify the gait observations discussed above. A thorough search of the relevant literature yielded that almost little to no work was done to detect neurological disorders based on the gait videos. There are works with the same goal but using motion sensor data, a time-consuming, expensive, and sophisticated approach. The sensor-data based approach is not pragmatic because we aim to potentially detect such disorders even outside hospitals, for public assistance. The predictors designed should be robust enough so that they do not depend on irregularities such as video-duration, background variation \cite{jerripothula2015qcce}, foreground variations ~\cite{anusha2020clothing,jerripothula2018quality,7025663}, and lighting conditions.

We address the above-described challenges in the following manner:
We create a dataset named \emph{NeuroSynGait} dataset, which is a synthetic dataset of videos\cite{jerripothula2018efficient} of normal human beings mimicking the four gait abnormalities discussed and demonstrating their normal gait. Besides, we develop six gait based features, namely, limb straightness, hand-leg co-ordination, upper-body straightness, body straightness, central distances, and mutual distances, for a frame. These features are summarized over the video using statistics like mean and variance for developing a video-level feature. We use landmarks (key-points) of the human pose for developing these features. The pose is estimated using AlphaPose \cite{fang2017rmpe, li2018crowdpose, xiu2018poseflow} technique, which already accounts for challenges posed by the background, foreground variations, lighting conditions, etc.
The derived video-level features are used to build classification models (using machine learning algorithms) to detect a gait abnormality, and possibly the related neurological disorder.  

The contributions made through this research are as follows: (i) A benchmark dataset named NeuroSynGait video dataset for detecting gait abnormalities has been proposed. (ii) Six gait features have been developed while focusing on the gait abnormalities caused by four different neurological disorders, namely Parkinson's disease, Diplegia, Hemiplegia, and Huntington's Chorea. (iii) We successfully build machine learning models to detect these gait abnormalities. 

\section{Related Works}
Early works on detecting neurological disorders using gait analysis started with \cite{lee2000huma} and \cite{piecha2007gait}. \cite{lee2000huma} proposed wearing a specific suit and having a constrained environment (like a plain background) to extract relevant gait features effectively and detect neurological disorders, \cite{piecha2007gait} proposed detection of neurological disorders from load distribution on foot. \cite{lee2000huma} easily segmented the body (thanks to the suit and plain background) and extracted a skeleton through thinning operation to extract the skeleton required for developing the relevant features.   Similarly, \cite{abdul2010sensor} proposed a sensor system to detect various vertical movements and, subsequently, neurological disorders. Thus, all these works are heavily constrained either by the arrangements required or the instruments required. Also, given the idea presented in \cite{piecha2007gait}, \cite{papavasileiou2017classification} collected data from smart-shoes to predict neurological disorders, specifically stroke and Parkinson's disease. Similarly, \cite{tunca2017inertial} used inertial sensors to extract relevant features and detect neurological disorders.

It is clear from above that gait analysis has been one of the prominent characteristics in detecting neurological disorders. However, to the best of our knowledge, a pose-estimation based approach has not been explored yet to perform gait analysis that particularly focuses on neurological disorders.         

\section{Proposed Method}
In this section, we describe what data we extract from the video, how we design our gait features (focusing on neurological disorders) at frame-level using such data, and how we derive video-level features from those frame-level features.  

\begin{table}
    \centering
      \caption{Notations for the key-points in the Human Pose}
    \begin{tabular}{|l|c|c|}
    \hline
         Key-point & Notation as set $P$'s element & Symbolic Notation \\
         \hline
         \hline
         Left Ear & $P_1$&$R_l$ \\
         Right Ear & $P_2$&$R_r$\\
         Left Shoulder & $P_3$&$S_l$\\
         Right Shoulder & $P_4$&$S_r$\\
         Left Elbow & $P_5$&$E_l$\\
         Right Elbow & $P_6$&$E_r$\\
         Left Wrist & $P_7$&$W_l$\\
         Right Wrist & $P_8$&$W_r$\\
         Left Hip & $P_9$&$H_l$\\
         Right Hip & $P_{10}$&$H_r$\\
         Left Knee & $P_{11}$&$K_l$\\
         Right Knee & $P_{12}$&$K_r$\\
         Left Ankle & $P_{13}$&$A_l$\\
         Right Ankle & $P_{14}$&$A_r$\\
         \hline
    \end{tabular}
  
    \label{tab:notation}
    \end{table}

\subsection{Data Extraction} 
We extract key-points of human pose using the AlphaPose technique \cite{fang2017rmpe}, which essentially provides a skeleton~\cite{jerripothula2017object,gao20182d} along with the coordinates of different key-points (or joints). We select only the key-points we need for our gait analysis, and those key-points have been mentioned in Table~\ref{tab:notation}.
Let the selected set of key-points of a human pose be denoted as $P=\{P_i|i=1,\cdots,|P|\}$. Also, let those key-points be denoted by individual symbols, as shown in Table~\ref{tab:notation}, for readability. Note that for ensuring brevity and readability of our equations, we denote the key-points either as elements of $P$ or using symbolic notations. Each key-point has two values: an x-coordinate value and a y-coordinate value of the key-point location on a human body. In summary, taking an example, we denote x-coordinate of the left ear ($R_l$) as $R_l(x)$ (or $P_1(x)$) and its y-coordinate as $R_r(y)$ (or $P_2(y)$).

\subsection{Limb Straightness} 
 \begin{figure}
        \begin{center}
            
            \includegraphics[width=0.47\linewidth,height=2.4cm]{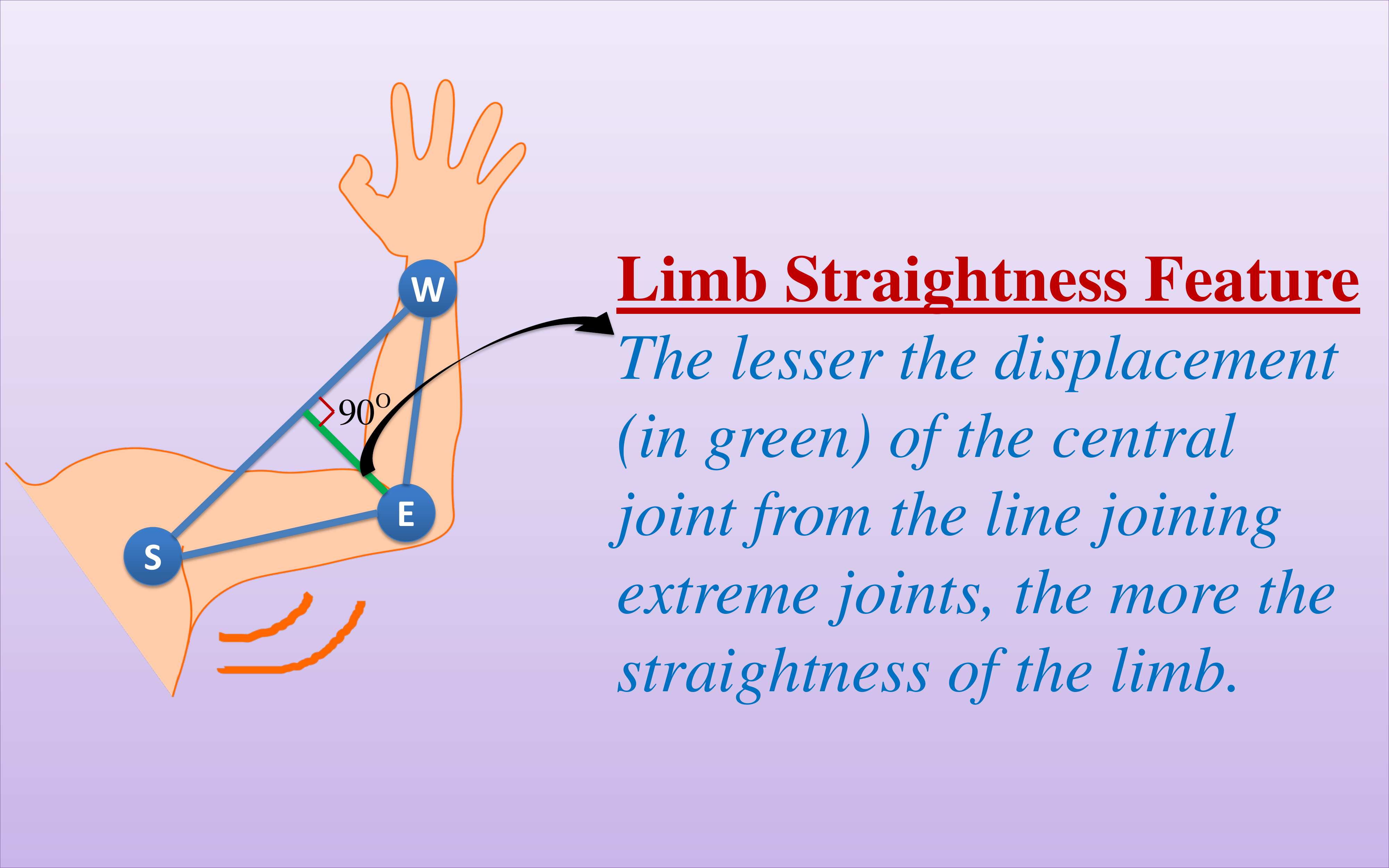}
            \includegraphics[width=0.47\linewidth,height=2.4cm]{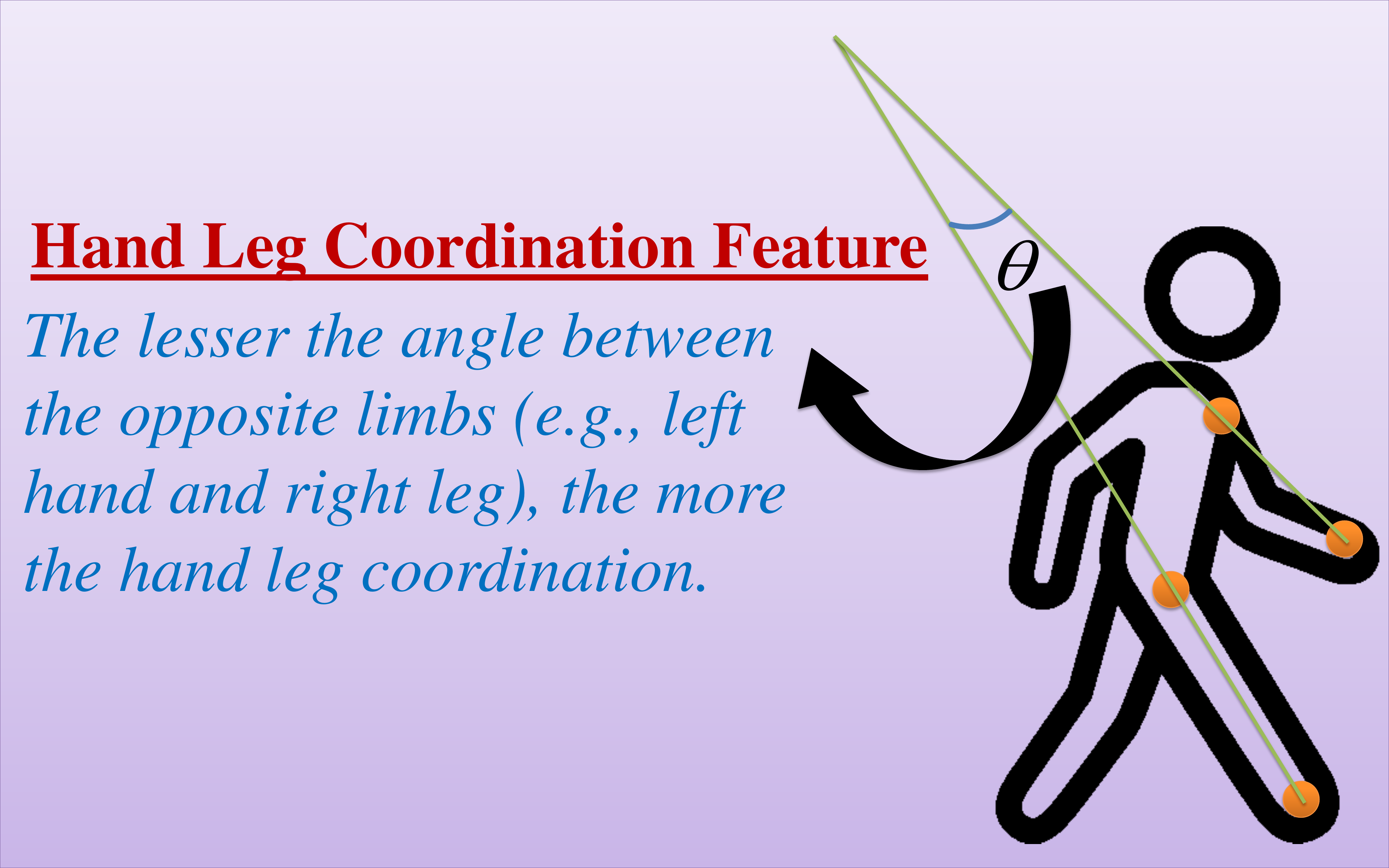}
        \end{center}
        \vspace{-4mm}
        \caption{Limb Straightness and Hand-leg Coordination}\label{fig:ls}
    \end{figure}

In the parkinsonian gait, we observe that the hands of the subject get bent almost like `V' shape while walking. Also, in choreiform gait, involuntary and unbalanced motion sometimes leads to a bending of hands and legs. Hence, we can exploit this observation to form and define our first feature \cite{7484309,9153789} named limb straightness. There are four limbs in total: two hands and two legs. While we can consider shoulder, elbow, and wrist landmark points present on a hand to measure its straightness, we can consider hip, knee, and ankle landmark points to do the same for legs. In Fig.~\ref{fig:ls}, we demonstrate how we compute the straightness of a hand limb. To quantify the observation, we measure the displacement of the central landmark point (elbow) from the line joining the other two landmark points (shoulder and wrist), which are extreme landmark points of the limb. We derive this distance in the following way, say for the left hand: (i) We compute the slope \Big(denoted as $m(S_l, W_l)$\Big) of the line joining $S_l$ and $W_l$. (ii) We compute the y-intercept \Big(denoted as $c(S_l,W_l)$\Big) of the line joining $S_l$ and $W_l$. (iii) We use central landmark point ($E_l$) coordinate values and slope calculated to obtain the perpendicular distance using the perpendicular distance of a point from a line formula. We use the following equations to arrive at our left-hand limb straightness \Big($LS(S_l, E_l, W_l)$\Big) feature:

\begin{align} 
m(S_l,W_l)&=\frac{S_l(y)-W_l(y)}{S_l(x)-W_l(x)}\\
c(S_l,W_l)&=\frac{W_l(y)S_l(x)-W_l(x)S_l(y)}{S_l(x)-W_l(x)}\\
LS(S_l,E_l,W_l)&=\frac{|m(S_l,W_l) E_l(x)+c(S_l,W_l)-E_l(y)|}{\sqrt{(m(S_l,W_l)^2+1}}.
\end{align} 
\begin{sloppypar}In the same way, we can find other limb straightness feature values also, which are $LS(S_r,E_r,W_r)$, $LS(H_l,K_l,A_l)$, and $LS(H_r,K_r,A_r)$. While the measured distance should be small normally, it would be much larger in the case of abnormality. Note that limb-bending can happen to even normal humans while holding something or walking on an uneven path, but such instances occur quite rarely compare to the occurrence of straight limbs. Therefore, we can know whether a person has abnormal gait or not only at a video level, after evaluating all the frames.
\end{sloppypar}

\subsection{Hand-leg Coordination}
We observe that while walking normally opposite pairs of an upper limb and a lower limb often tend to swing parallel to each other, i.e., left-hand swings parallelly with the right-leg, and right-hand swings parallelly with left-leg. However, we note that when a person affected with Hemiplegia or Diplegia walks, our observation of parallelism gets disrupted due to the abnormal conical movement of legs. Such disruption occurs in the case of Parkinson's Disease as well due to bent-hands and almost straight legs. So, with the motivation of encoding the hand-leg co-ordination using parallelism of pairs of opposite limbs, we compute the angle between the hand and the leg present in those pairs, as shown in Fig.~\ref{fig:ls}. We derive this angle in the following way, say for the pair comprising left-hand, and right-leg limbs: (i) We compute the slope of the hand \Big($m(S_l, W_l)$\Big). (ii) We compute the slope of the leg \Big($m(H_r, A_r)$\Big). (iii) We compute our hand-leg co-ordination feature \Big($HL(S_l, W_l, H_r, A_r)$\Big) for the pair by finding the difference between the corresponding angles of two slopes computed, as shown below:
\begin{equation}
HL(S_l,W_l,H_r,A_r)=|tan^{-1}\Big(m(S_l,W_l)\Big)-tan^{-1}\Big(m(S_l,W_l)\Big)|
\end{equation}

This angle turns out to be smaller in the normal case and possesses quite a significant value in the abnormal case. We can compute this angle for the other pair as well, i.e., involving right-hand and left-leg, i.e., $HL(S_r, W_r, H_l, A_l)$. In this way, there are two hand-leg co-ordination feature values for a frame. Like the limb straightness feature, this feature can also be analyzed at the video-level only to arrive at a proper decision.

\subsection{Upper Body Straightness}
\begin{figure}
        \begin{center}            
            \includegraphics[width=0.47\linewidth,height=3cm]{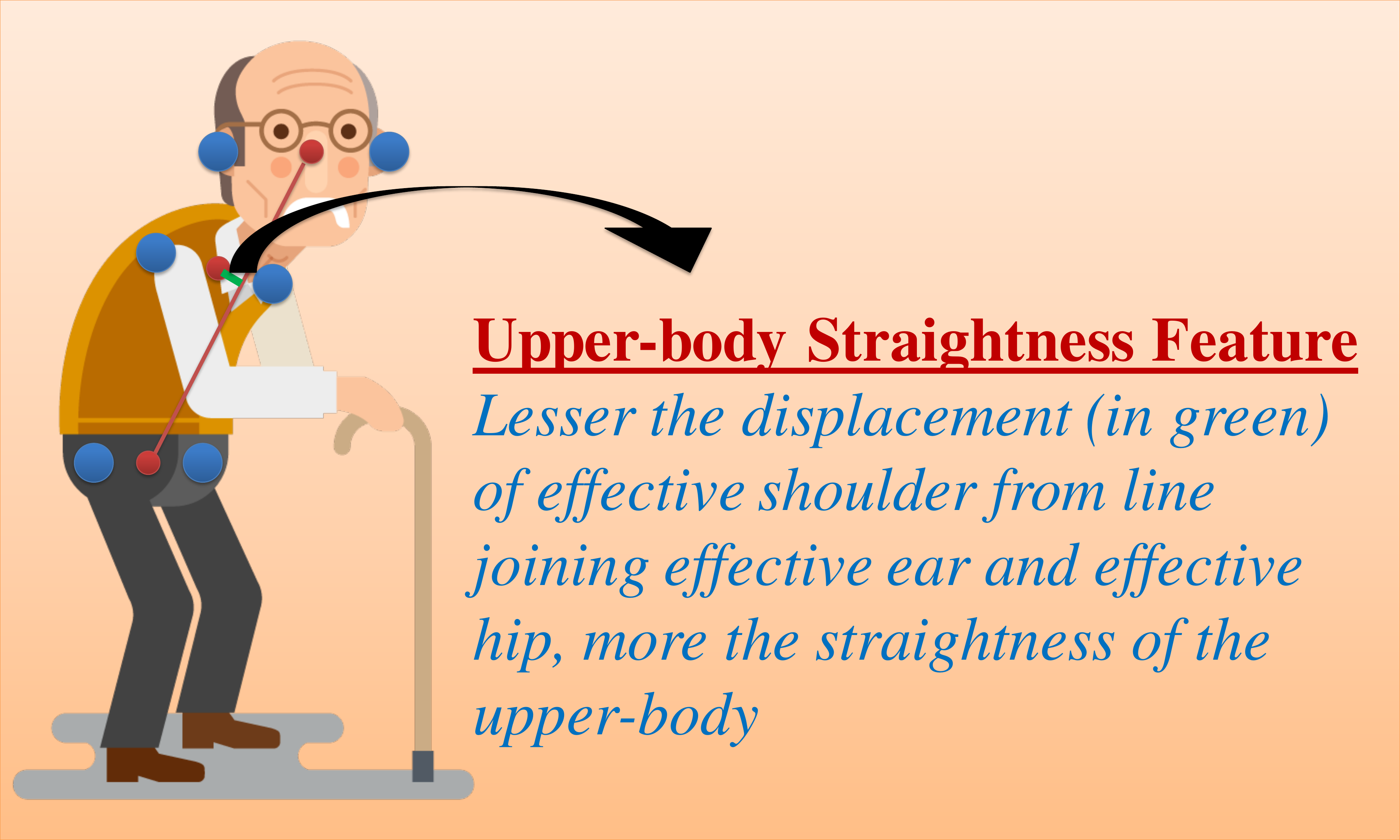}
            \includegraphics[width=0.47\linewidth,height=3cm]{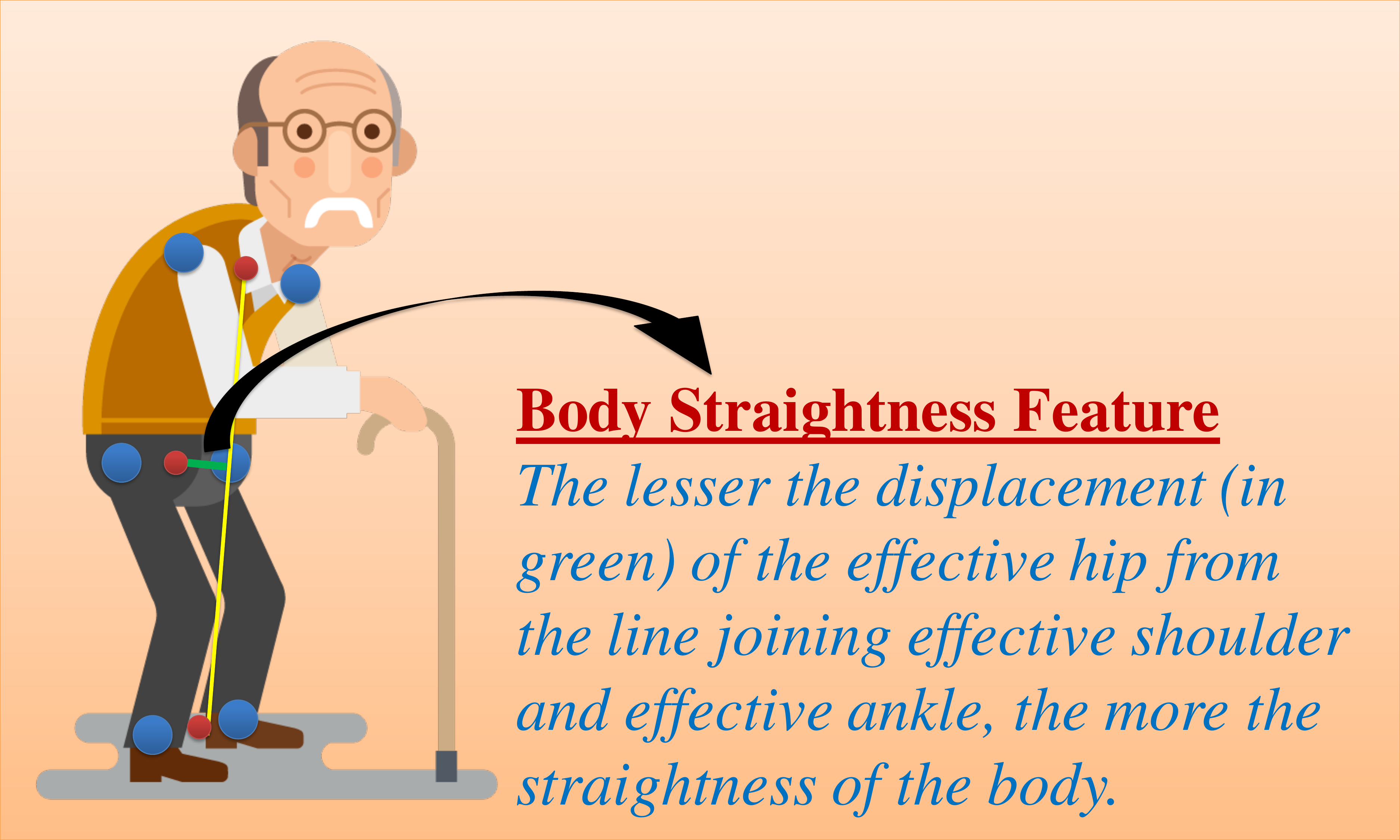}
        \end{center}
       \vspace{-4mm}
        \caption{Upper-body and Total-body Straightness }\label{fig:us}
    \end{figure}
The observation that upper-body bends in the parkinsonian and diplegic gaits leads us to our next gait feature: upper body straightness.This straightness particularly gets affected in choreiform gait with the unstable neck. To quantify this observation, we define our upper-body straightness in terms of displacement of effective shoulder coordinates from the line joining the effective hip and effective ear, as shown in the Fig.~\ref{fig:us}. By effective shoulder/hip/ear, we mean midpoint of the two ears/shoulders/hips. Ideally, the effective ear, effective shoulder, and effective hip should be collinear. We measure this non-collinearity in the following manner: Similar to limb straightness, we compute this feature by computing the perpendicular distance of effective shoulder from the line joining the effective hip and the effective ear. The final expression of our upper-body straightness (US) feature turns out to be the following:
\begingroup\makeatletter\def\f@size{7}\check@mathfonts
\def\maketag@@@#1{\hbox{\m@th\large\normalfont#1}}%
\begin{multline}
US=0.5\Bigg|\frac{\frac{R_l(y)+R_r(y)-H_l(y)-H_r(y)}{R_l(x)+R_r(x)-H_l(x)-H_r(x)}\Big(S_l(x)+S_r(x)\Big)}{\sqrt{1+(\frac{R_l(y)+R_r(y)-H_l(y)-H_r(y)}{R_l(x)+R_r(x)-H_l(x)-H_r(x)})^2}}\\+ \frac{\Big(R_l(x)+R_r(x)\Big)\Big(H_l(y)+H_r(y)\Big)-\Big(H_l(x)+H_r(x)\Big)\Big(R_l(y)+R_r(y)\Big)}{\Big(R_l(x)+R_r(x)-H_l(x)-H_r(x)\Big)\sqrt{1+(\frac{R_l(y)+R_r(y)-H_l(y)-H_r(y)}{R_l(x)+R_r(x)-H_l(x)-H_r(x)})^2}}\\- \frac{\Big(S_l(y)+S_r(y)\Big)}{\sqrt{1+(\frac{R_l(y)+R_r(y)-H_l(y)-H_r(y)}{R_l(x)+R_r(x)-H_l(x)-H_r(x)})^2}}\Bigg|.
\end{multline}
\endgroup
So, the lesser the displacement of the effective shoulder from the line joining effective ear and effective hip, the more the straightness of the upper-body. Again, this feature also can be analysed only at video-level for a conclusive decision making, for even a healthy human may bend forward sometimes.

\subsection{Body Straightness}

During hemiplegic gait, since one of the legs follow a conical motion while walking, the entire body can not be in a straight form. Similarly, with all kinds of arbitary movements at different joints of the body in the Choreiform gait, we can't expect body to be straight in such a scenario either. Same goes with the parkisonian gait where one's upper body comes forward quite characteristically. Therefore, the next feature is named as body straightness. This measures the straightness of the entire body as a whole. For the same, the displacement of the effective hip is computed from the line joining the effective shoulder and effective ankle joints, as shown in the Fig.~\ref{fig:us}. Similar to upper-body straightness, the body straightness (BS) value is computed as following, considering shoulders, hips, and ankles positions:
\begingroup\makeatletter\def\f@size{7}\check@mathfonts
\def\maketag@@@#1{\hbox{\m@th\large\normalfont#1}}%
\begin{multline}
BS=0.5\Bigg|\frac{\frac{S_l(y)+S_r(y)-A_l(y)-A_r(y)}{S_l(x)+S_r(x)-A_l(x)-A_r(x)}\Big(H_l(x)+H_r(x)\Big)}{\sqrt{1+(\frac{S_l(y)+S_r(y)-A_l(y)-A_r(y)}{S_l(x)+S_r(x)-A_l(x)-A_r(x)})^2}} \\+ \frac{\Big(S_l(x)+S_r(x)\Big)\Big(A_l(y)+A_r(y)\Big)-\Big(A_l(x)+A_r(x)\Big)\Big(S_l(y)+S_r(y)\Big)}{\Big(S_l(x)+S_r(x)-A_l(x)-A_r(x)\Big)\sqrt{1+(\frac{S_l(y)+S_r(y)-A_l(y)-A_r(y)}{S_l(x)+S_r(x)-A_l(x)-A_r(x)})^2}}\\- \frac{\Big(H_l(y)+H_r(y)\Big)}{\sqrt{1+(\frac{S_l(y)+S_r(y)-A_l(y)-A_r(y)}{S_l(x)+S_r(x)-A_l(x)-A_r(x)})^2}}\Bigg|.
\end{multline}
\endgroup
So, the lesser the displacement of the effective hip from the line joining effective shoulder and effective ankle, the more the straightness of the body. Again, this feature also needs to be analyzed at video-level for effective results.

\subsection{Central Distances}

\begin{figure}
        \begin{center}
            
            \includegraphics[width=0.53\linewidth,height=3cm]{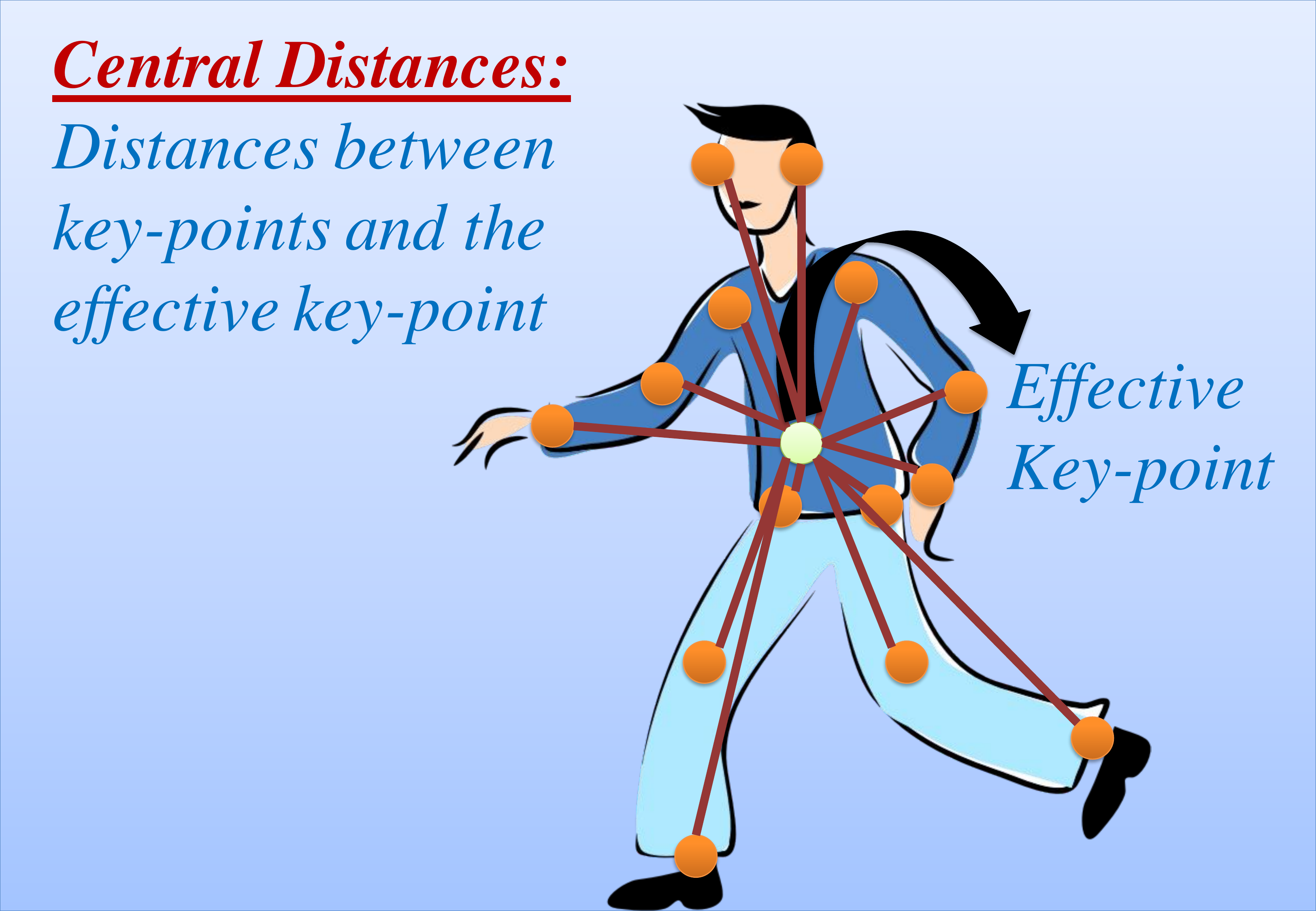}
            \includegraphics[width=0.32\linewidth,height=3cm]{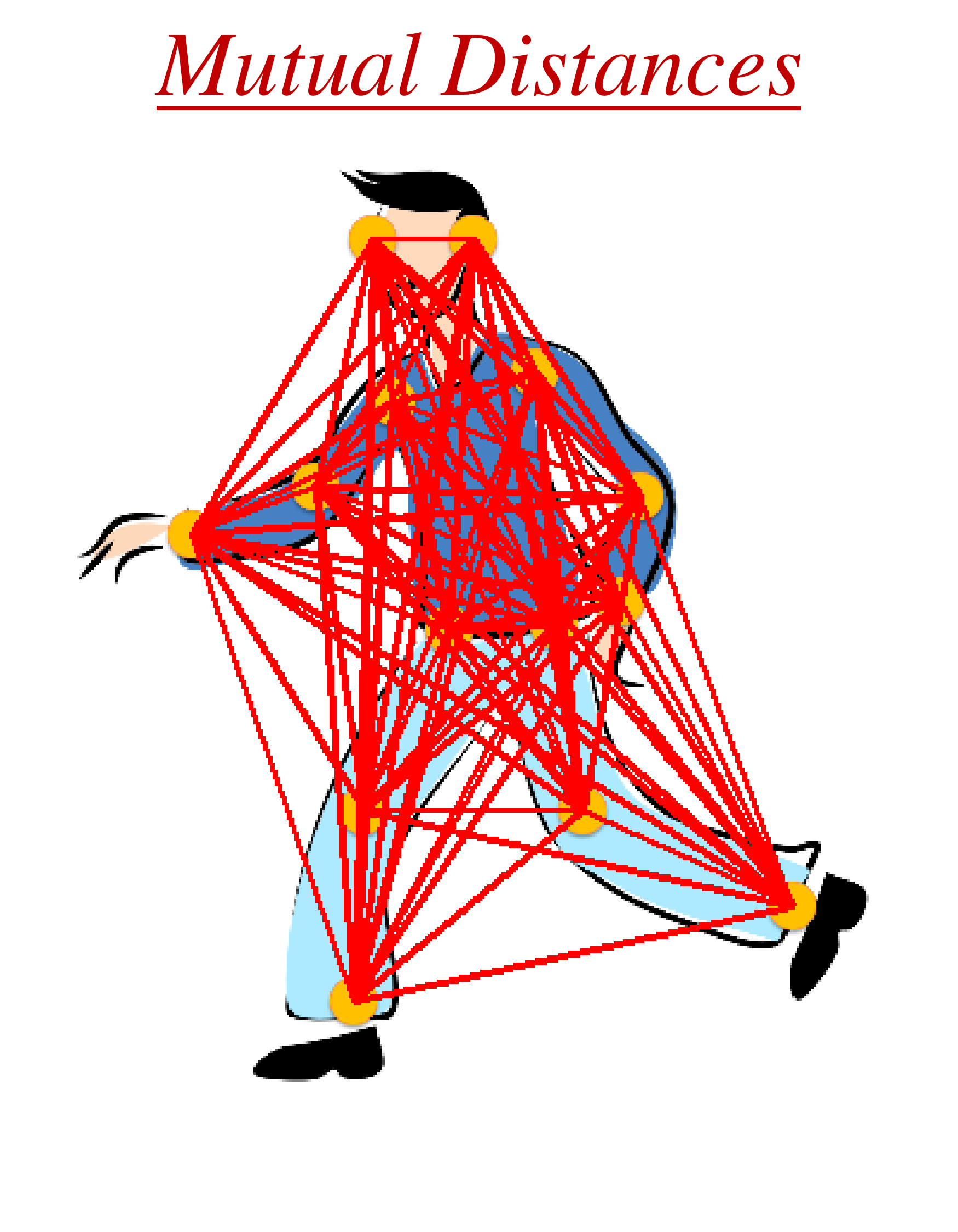}
        \end{center}
        \vspace{-4mm}
        \caption{Central Distances and Mutual Distances}\label{fig:cd}
    \end{figure}
We develop a feature named central distances to capture the distances of all the landmark points with what we call as effective key-point, as shown in Fig.\ref{fig:cd}. Such distances can serve interesting information to detect any anomalies in the gait of a person. Particularly for choreiform gait, these distances keep changing due to the kind of jerks involved in it. We define effective key-point as the centroid of all the landmark points. Let the central distance of a landmark point $P_i$ be denoted as CD(i). In this way, we have all $|P|=14$ distances per frame to serve as a piece of potential information to detect a person's anomalous gait. Note that there may be a person's size-variations in the video, which can affect the uniformity in the ranges of these distances across the frames. To account for this, we normalize these distances by dividing them with the maximum distance obtained.

\subsection{Mutual Distances}

Similar to central distances, we develop another feature named mutual distances to capture the distances of different landmark points with each other, as shown in Fig.\ref{fig:cd}. Such distances can also serve as other interesting information to detect any anomalies in a person's gait. Let distance between $P_i$ and $P_j$ be denoted $MD(i,j)$. In this way, we can have all the $^{|P|}C_2=91$ mutual distances per frame to serve as another piece of information to detect the abnormal gait of a person potentially. Note that, similar to central distances, here also, we normalize these distances by dividing them with the maximum distances obtained. 

\subsection{Video-level Features}
The features discussed so far were all at frame-level, capturing important information about a person's pose in a particular frame. Their details are given in Table~\ref{tab:feature}. Note that abnormal poses in gait can be demonstrated by even healthy persons and persons once in a while. So, real patients are identifiable only over a while. We currently have a 4+2+1+1+14+91=113 dimensions feature vector at frame-level. To summarize these feature vectors at video level, we use statistical tools such as mean and standard deviation of these feature vectors across the frames in a video and concatenate them. In this way, we obtain a 113x2=226 dimensions feature vector at the video-level.

\begin{figure}
        \begin{center}
            
            \includegraphics[width=1\linewidth]{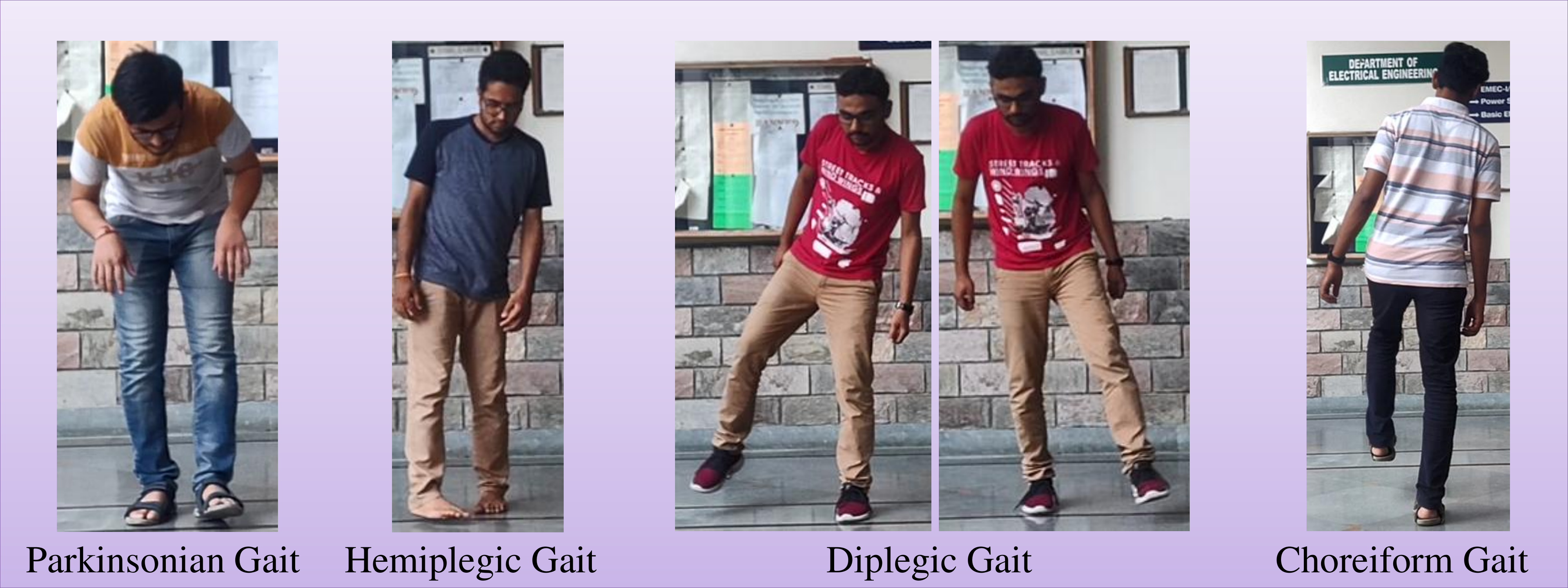}
        \end{center}
        \vspace{-1mm}
        \caption{A glimpse of our dataset as we can see our participants mimicking different abnormal gaits}\label{fig:data}
    \end{figure}
\begin{table}
\centering
\caption{Details of our features for a frame}\label{tab:feature}
\vspace{-1mm}
\resizebox{0.5\textwidth}{!}{\begin{tabular}{|l|c|c|}
\hline
Feature & Notations & Dimensions \\ \hline
Limb Straightness     &   $\{LS(S_l,E_l,W_l),LS(S_r,E_r,W_r),$                &            \\      &   $LS(H_l,K_l,A_l),LS(H_r,K_r,A_r)\}$                & 4           \\ \hline
Hand-Leg Coordination       &        $\{HL(S_l,W_l,H_r,A_r),$           &        \\
       &        $HL(S_r,W_r,H_l,A_l)\}$           & 2       \\ \hline
Upper-body Straightness &      $\{US\}$            & 1  \\ \hline
Body Straightness &      $\{BS\}$            & 1  \\ \hline
 Central Distances     & $\{CD(i)|i=1,\cdots,|P|\}$               & 14                     \\ \hline
  Mutual Distances     &    $\{MD(i,j)|i,j=1,\cdots,|P|\}$               & 91                        \\ \hline
\end{tabular}}
\vspace{2mm}
\centering
\caption{Description of our NeuroSynGait dataset}\label{tab:data}
\vspace{-1mm}
\resizebox{0.5\textwidth}{!}{
\begin{tabular}{|l|c|c||c|}
\hline
Abnormality & No. of Training Videos & No. of Testing Videos & Total Videos \\ \hline
Chorieform     & 38                  & 13                 & 51           \\ \hline
Diplegia       & 41                  & 14                 & 55           \\ \hline
Hemiplegia     & 52                  & 18                 & 70           \\ \hline
Normal         & 23                  & 8                  & 31           \\ \hline
Parkinson      & 38                  & 13                 & 51           \\ \hline
\end{tabular}}
\end{table}
\begin{table*}[!htbp]
\centering
\caption{Cross-validation classification accuracy for multiple abnormalities detection using different methods and machine learning algorithms. }\label{tab:cvca}
\vspace{-1mm}
\resizebox{0.75\textwidth}{!}{%
\begin{tabular}{|l|c|c|c|c|c|c|c|c|c|}
\hline
\multicolumn{1}{|c|}{} & kNN   & Tree  & SVM   & SGD            & \begin{tabular}[c]{@{}c@{}}Random\\ Forest\end{tabular} & \begin{tabular}[c]{@{}c@{}}Neural\\ Network\end{tabular} & \begin{tabular}[c]{@{}c@{}}Naive\\ Bayes\end{tabular} & \begin{tabular}[c]{@{}c@{}}Logistic\\ Regression\end{tabular} & AdaBoost \\ \hline
AlphaPose\cite{fang2017rmpe}                 & 0.260 & 0.458 & 0.339 & 0.547          & 0.505                                                   & 0.490                                                    & 0.307                                                 & 0.656                                                & 0.469    \\
 \hline
3D-CNN\cite{tran2015learning}             & 0.260 & 0.333 & 0.432 & 0.568          & 0.344                                                   & 0.495                                                    & 0.318                                                 & 0.573                                                & 0.375    \\ \hline
AlphaPose\cite{fang2017rmpe}+&&&&&&&&&
\\
3D-CNN\cite{tran2015learning}        & 0.479 & 0.396 & 0.531 & 0.635          & 0.453                                                   & 0.609                                                    & 0.469                                                 & \textcolor{blue}{\textbf{0.667}}                                                & 0.365    \\
 \hline\hline
Ours           & \textcolor{blue}{\textbf{0.557}} & \textcolor{blue}{\textbf{0.682}} & \textcolor{blue}{\textbf{0.820}}  & \textcolor{blue}{\textbf{0.818}} & \textcolor{blue}{\textbf{0.714}}                                                   & \textcolor{blue}{\textbf{0.797}}                                                    & \textcolor{blue}{\textbf{0.661}}                                                 & 0.276                                                         & \textcolor{blue}{\textbf{0.641}}    \\ \hline
\end{tabular}}
\vspace{2mm}
\centering
\caption{Test classification accuracies for multiple abnormalities detection using different methods and machine learning algorithms.}\label{tab:tca}
\vspace{-1mm}
\resizebox{0.75\textwidth}{!}{%
\begin{tabular}{|l|c|c|c|c|c|c|c|c|c|}
\hline
\multicolumn{1}{|c|}{} & kNN   & Tree  & SVM   & SGD            & \begin{tabular}[c]{@{}c@{}}Random\\ Forest\end{tabular} & \begin{tabular}[c]{@{}c@{}}Neural\\ Network\end{tabular} & \begin{tabular}[c]{@{}c@{}}Naive\\ Bayes\end{tabular} & \begin{tabular}[c]{@{}c@{}}Logistic\\ Regression\end{tabular} & AdaBoost \\ \hline
AlphaPose\cite{fang2017rmpe}                 & 0.197 & 0.591 & 0.470 & 0.515          & 0.470                                                   & 0.561                                                    & 0.455                                                 & \textcolor{blue}{\textbf{0.758}}                                                & 0.576    \\ \hline
3D-CNN\cite{tran2015learning}             & 0.273 & 0.379 & 0.455 & 0.530          & 0.379                                                   & 0.485                                                    & 0.288                                                 & 0.591                                                & 0.348    \\ \hline
AlphaPose\cite{fang2017rmpe}+&&&&&&&&&
\\
3D-CNN\cite{tran2015learning}         & 0.424 & 0.394 & 0.545 & 0.606          & 0.379                                                   & 0.636                                                    & 0.470                                                 & 0.652                                                & 0.379    \\ \hline\hline
Ours           & \textcolor{blue}{\textbf{0.636}} & \textcolor{blue}{\textbf{0.636}} & \textcolor{blue}{\textbf{0.788}} & \textcolor{blue}{\textbf{0.864}} & \textcolor{blue}{\textbf{0.833}}                                                   & \textcolor{blue}{\textbf{0.788}}                                                    & \textcolor{blue}{\textbf{0.773}}                                                 & 0.212                                                         & \textcolor{blue}{\textbf{0.712}}    \\ \hline

\end{tabular}}
\end{table*}

\begin{table}[!htbp]
\centering
\caption{The details of the best models for Parkinsonian gait detection using different methods across different machine learning algorithms. By best, we mean sum of cross validation and test classification accuracies are highest.}\label{tab:iip}
\vspace{-1mm}
\resizebox{0.5\textwidth}{!}{\begin{tabular}{|l|c|c|c||c|}
\hline
Details & AlphaPose\cite{fang2017rmpe} & 3D-CNN\cite{tran2015learning} & AlphaPose\cite{fang2017rmpe}+3D-CNN\cite{tran2015learning}& Ours\\ \hline
Cross Validation   & \textcolor{blue}{\textbf{0.984}} & 0.918 & 0.967 & \textcolor{blue}{\textbf{0.984}} \\\hline
Test & \textcolor{blue}{\textbf{1.000}} & 0.905 & 1.000 & \textcolor{blue}{\textbf{1.000}} \\\hline
Algorithm & NN    & LR    & SVM   & RF    \\\hline

\end{tabular}}

\vspace{2mm}
\centering
\caption{The details of the best models for Hemiplegic gait detection using different methods across different machine learning algorithms. By best, we mean sum of cross validation and test classification accuracies are highest.}\label{tab:iih}
\vspace{-1mm}
\resizebox{0.5\textwidth}{!}{\begin{tabular}{|l|c|c|c||c|}
\hline
Details & AlphaPose\cite{fang2017rmpe} & 3D-CNN\cite{tran2015learning} & AlphaPose\cite{fang2017rmpe}+3D-CNN\cite{tran2015learning}& Ours\\ \hline
Cross Validation   & 0.867 & 0.827 & 0.800 & \textcolor{blue}{\textbf{0.893}} \\\hline
Test & \textcolor{blue}{\textbf{0.962}} & 0.846 & 0.769 & 0.960 \\\hline
Algorithm & LR    & LR    & LR    & NN    \\\hline

\end{tabular}}
\end{table}

\section{Experimental Results}
In this section, we first discuss the dataset developed by us, named NeuroSynGait video dataset. Second, we discuss the details of the different experiments conducted. Third, we discuss the different results we have obtained. 

\subsection{Dataset}

Since neurological disorders are rare, we show different videos of the patients' gait with these disorders to healthy individuals and request them to mimic those gait abnormalities. To mimic those videos properly, we also request them to read the publicly available descriptions of each gait\footnote{https://neurologicexam.med.utah.edu/adult/html/gait\_abnormal.html} before they mimic these gaits. In this way, we obtain a substantial number of videos of gait abnormalities. The details of these videos are given in the Table~\ref{tab:data}. We collected 258 videos and labeled them as either one of the four gait abnormality or as normal. In this way, our dataset has a total of 5 classes. We adopt the 3:1 ratio for distributing the collected videos into training and testing subsets. A glimpse of participants performing the abnormal gait is given in Fig.~\ref{fig:data}.

\subsection{Experiment Details}
We use various machine learning algorithms such as Ada Boost (AB) \cite{freund1997decision}, Decision Tree (DT) \cite{quinlan1986induction}, k-Nearest Neighbors (kNN) \cite{altman1992introduction}, Logistic Regression (LR) \cite{walker1967estimation}, Naive Bayes (NB) \cite{maron1961automatic}, Neural Networks (NN) \cite{mcculloch1943logical}, Random Forests (RF) \cite{ho1995random}, Stochastic Gradient Descent (SGD) \cite{robbins1951stochastic}, Support Vectors Machine (SVM) \cite{cortes1995support} to learn models that can predict the presence of a particular disorder. We report the classification accuracies of both cross-validation phase and testing phase. 

We conduct two sets of experiments: multiple abnormality prediction and individual abnormality prediction. In multiple abnormality prediction, there are five classes (4 abnormalities and 1 normal). In individual abnormality prediction, there are only two classes: the concerned abnormality and the normal. We compare our method with existing 3D-CNN~\cite{tran2015learning} and AlphaPose~\cite{fang2017rmpe}. While \cite{tran2015learning} proposed spatiotemporal features using deep 3D ConvNets, \cite{fang2017rmpe} proposed a regional multi-person pose estimation framework, which we employ for developing our features. We compare with both the techniques and their combination in the following manner. (i) AlphaPose~\cite{fang2017rmpe}, where we use landmark points as features for a video-frame and summarize them at video-level just like we do. (ii) 3D-CNN~\cite{tran2015learning}, where features obtained for a set of frames are summarized at video-level just like we do. (iii) AlphaPose~\cite{fang2017rmpe} + 3D-CNN~\cite{tran2015learning}, where we extract video features using \cite{tran2015learning} on the skeleton videos generated by \cite{fang2017rmpe}.

\begin{table}
\centering
\caption{The details of the best models for Diplegic gait detection using different methods across different machine learning algorithms. By best, we mean the sum of cross-validation and test classification accuracies are highest.}\label{tab:iid}
\vspace{-1mm}
\resizebox{0.5\textwidth}{!}{\begin{tabular}{|l|c|c|c||c|}
\hline
Details & AlphaPose\cite{fang2017rmpe} & 3D-CNN\cite{tran2015learning} & AlphaPose\cite{fang2017rmpe}+3D-CNN\cite{tran2015learning}& Ours\\ \hline
Cross Validation   & \textcolor{blue}{\textbf{0.891}} & 0.719 & 0.781 & 0.734 \\\hline
Test & 0.818 & 0.909 & 0.818 & \textcolor{blue}{\textbf{0.955}} \\\hline
Algorithm & LR    & LR    & NN    & RF    \\\hline

\end{tabular}}
\vspace{2mm}
\centering
\caption{The details of the best models for Choreiform gait detection using different methods across different machine learning algorithms. By best, we mean sum of cross validation and test classification accuracies are highest.}\label{tab:iic}
\vspace{-1mm}
\resizebox{0.5\textwidth}{!}{\begin{tabular}{|l|c|c|c||c|}
\hline
Details & AlphaPose\cite{fang2017rmpe} & 3D-CNN\cite{tran2015learning} & AlphaPose\cite{fang2017rmpe}+3D-CNN\cite{tran2015learning}& Ours\\ \hline
Cross Validation   & 0.689 & 0.754 & 0.721 & \textcolor{blue}{\textbf{0.951}} \\\hline
Test & 0.905 & \textcolor{blue}{\textbf{0.952}}  & 0.762 & \textcolor{blue}{\textbf{0.952}} \\\hline
Algorithm & RF    & LR    & NN    & NN    \\\hline

\end{tabular}}
\end{table}

\subsection{Results}
From Tables~\ref{tab:cvca}-\ref{tab:tca}, it's clear that, in the case of multiple abnormality prediction, our features demonstrate superior performance compared to all the three existing ones, as it outperforms them in 8/9 learning algorithms, both at cross-validation and test stages. Our multiple abnormality detector is best learned using the SGD algorithm as its sum of cross-validation (81.8\%) and test accuracy (86.4\%) is found to be best. Following the same idea of what is best, we identify the best individual gait abnormality detectors for each method and report the results in Tables~\ref{tab:iip}-\ref{tab:iic}. Compared to other methods, our best individual gait abnormality detectors consistently scored more than 95\% in terms of test classification accuracy and achieved the highest cross-validation and test classification accuracies for 3/4 abnormalities.

\subsection*{Conclusion}
We develop several novel gait features, namely limb straightness, hand-leg co-ordination, upper-body straightness, body straightness, central distances, and mutual distances to detect gait abnormalities caused by neurological disorders in videos. We employ the key-points of the human pose for designing them. Our experiments demonstrate their superior performance in comparison to the existing ones.  

\bibliographystyle{IEEEtran}
\bibliography{main}

\end{document}